\documentclass{article} 
\usepackage{iclr2027_conference,times}

\usepackage{amsmath,amsfonts,bm}

\def\eqref#1{equation~\ref{#1}}

\def\1{\bm{1}}

\DeclareMathAlphabet{\mathsfit}{\encodingdefault}{\sfdefault}{m}{sl}
\SetMathAlphabet{\mathsfit}{bold}{\encodingdefault}{\sfdefault}{bx}{n}

\usepackage{hyperref}
\usepackage{url}
\usepackage{booktabs}
\usepackage{graphicx}
\usepackage{placeins}

\title{Effective Does Not Mean Useful: Conditional Functional Substitutability for Redundancy and Scaling in Transformers}

\author{
Jiaheng Chen$^{1,2,*}$,
Jiaxing Li$^{1,2}$,
Yucheng Xiao$^{1,2}$,
Xinyong Cai$^{4}$,
\\
Juncheng Bu$^{5}$,
Lan Yu$^{2}$,
Tinghe Zhang$^{3,*}$
\\[2mm]
$^{1}$Harbin Institute of Technology, Shenzhen, China \\
$^{2}$Northeastern University, Shenyang, China \\
$^{3}$Tsinghua University, Beijing, China \\
$^{4}$Tsinghua Shenzhen International Graduate School, Tsinghua University, Shenzhen, China \\
$^{5}$Peking University Shenzhen Graduate School, Shenzhen, China
\\[1mm]
$^{*}$Corresponding authors
}

\iclrfinalcopy 
\begin{document}

\maketitle

\fancyhead{}

\begin{abstract}
Modern neural networks scale predictably, yet the mechanisms behind these regularities remain unclear.
Neural redundancy is typically characterized by component importance or representational similarity, both indirect proxies.
We view redundancy as an input-conditioned, dynamic relation: intermediate computational states are functionally redundant when they induce similar downstream responses. We introduce Conditional Functional Substitutability (CFS) to directly characterize such functional substitution. 
CFS exposes functional relations and reduction potential missed by conventional importance- and similarity-based measures.
Across modalities and Transformer families, CFS reveals systematic functional reorganization with scale. Controlled scaling further shows that performance gains need not track growth in substitutability, while fixed-capacity models with more independent functional structure perform better, providing a functional account of diminishing returns.
Predicted CFS further enables dynamic computation with a better performance--computation trade-off than importance-based component selection, suggesting new directions for redundancy-aware computation and more efficient model scaling.
\end{abstract}

\section{Introduction}


Modern deep learning has advanced through scaling, with performance often following predictable empirical power laws with diminishing returns \citep{kaplan2020scaling,hoffmann2022compute}. Yet the microscopic mechanisms underlying these regularities remain poorly understood.

Neural networks also contain substantial redundancy~\citep{michel2019sixteen,voita2019analyzing,dalvi2020redundancy,bian2021attention}, typically characterized through component importance or representational similarity. Importance measures removal or perturbation effects, whereas similarity compares parameters, activations, or representations; both remain indirect proxies for redundancy. We argue that redundancy is more fundamentally an input-conditioned, dynamic relation between intermediate computational states: states that are distant in representation space may still be functionally redundant if they induce similar downstream responses.


We introduce Conditional Functional Substitutability (CFS) to measure whether one intermediate state can reproduce another's downstream response for a given input. CFS is input-conditioned and directional, revealing functional reduction potential missed by importance and similarity proxies: a computation may be locally effective without being independently indispensable.


With this functional view, we revisit scaling across vision and language Transformers. Across model families, scaling systematically reorganizes functional relations. Controlled scaling shows that performance gains can accompany markedly different growth in substitutability, while at fixed capacity more independent functional structure is associated with better language-modeling performance. Training dynamics further reveal how these relations form and consolidate. Together, these results provide a microscopic functional account of diminishing returns.

\begin{figure}[t]
    \centering
    \includegraphics[width=0.96\linewidth]{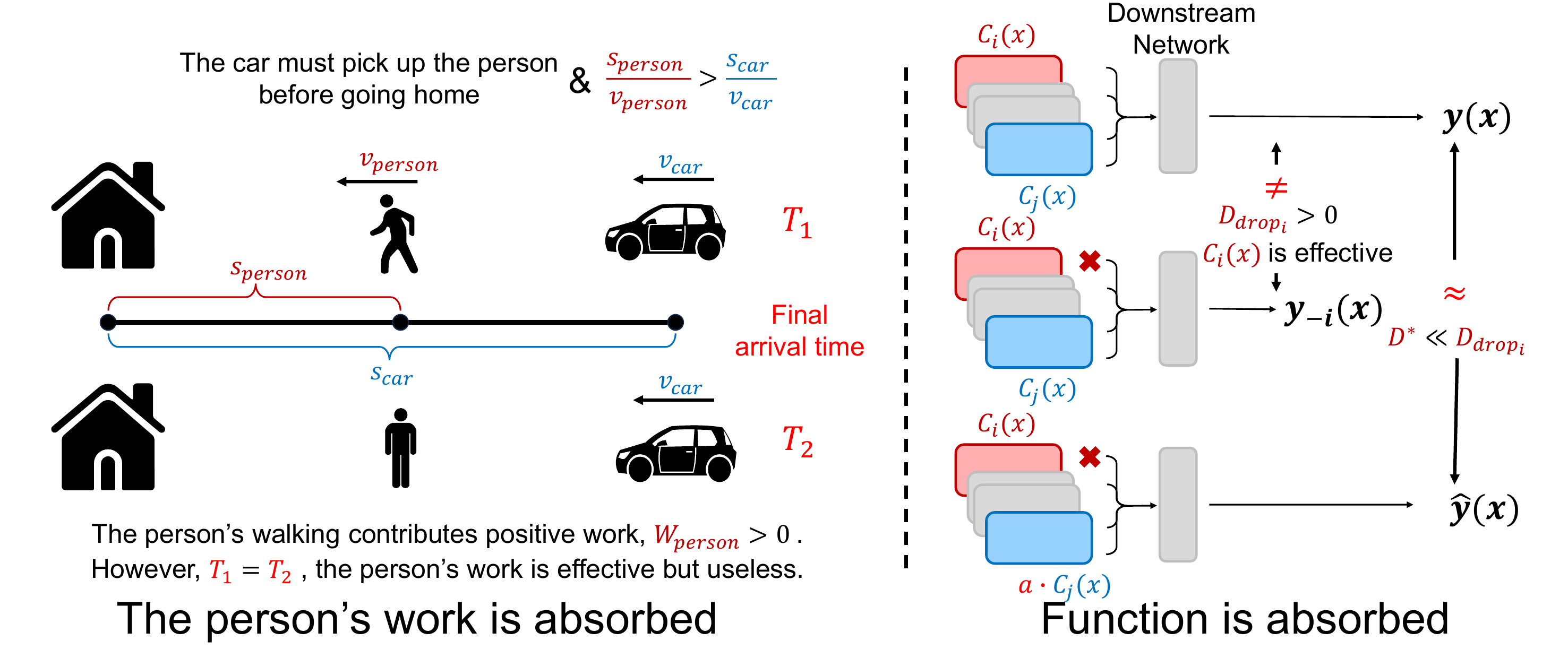}
    \vspace{-8pt}
    \caption{
    A locally effective computation may lack independent global utility when its downstream role is absorbed by an alternative path. CFS directly characterizes this functional substitutability.
    }
    \vspace{-4pt}
    \label{fig:intuition}
\end{figure}

Beyond analysis, we test whether CFS can guide computation by predicting it from forward model states and using the predictions for input-conditioned component selection. CFS-guided allocation achieves a better performance--computation trade-off than importance-based selection. More broadly, the link between scaling strategy and functional organization suggests that functional structure may itself be a design variable for improving scaling efficiency and potentially mitigating diminishing returns.

Our contributions are summarized as follows:

\begin{itemize}

\item We point out that existing importance- and similarity-based measures are fundamentally indirect proxies for neural redundancy. We introduce Conditional Functional Substitutability (CFS), which directly characterizes redundancy through input-conditioned, directional functional substitutability between computational components.

\item We provide the first systematic study of Transformer scaling from the perspective of functional redundancy. Across modalities, model families, controlled scaling axes, and training dynamics, we uncover consistent structural reorganization of functional computation with scale, providing a functional account of diminishing returns and linking microscopic functional organization to macroscopic scaling behavior.

\item Based on CFS, we establish oracle upper bounds on removable computation under joint interventions and develop a predictable dynamic computation mechanism that achieves a better performance--computation trade-off than importance-based selection, recovering part of the oracle advantage without intervention-time information.

\end{itemize}

\section{Related Work}



\paragraph{Neural redundancy and functional substitutability.}
Many Transformer heads or larger units can be removed with limited performance loss, motivating importance-, sensitivity-, and sparsity-based pruning
\citep{michel2019sixteen,voita2019analyzing,budhraja2020weak,ma2023llmpruner,men2024shortgpt},
while parameter, attention, and representation similarity have been used for compression or computation reuse
\citep{dalvi2020redundancy,bian2021attention,bhojanapalli2021reuse}.
Model stitching probes functional compatibility by testing whether one representation supports another model's downstream computation
\citep{bansal2021revisiting}.
Intervention-based interpretability localizes causal components and pathways
\citep{zhang2023patching}, while self-repair and Conditional Co-Ablation expose compensation and dormant backups after intervention
\citep{mcgrath2023hydra,rushing2024selfrepair,gong2026coax}.
CFS instead models replacement directly: whether one component's downstream effect can be recovered by another for a given input, yielding a pairwise, directional, and input-conditioned relation for functional redundancy and model-scale organization.


\paragraph{Neural scaling and its mechanisms.}
Scaling laws describe predictable performance changes with scale
\citep{kaplan2020scaling,hoffmann2022compute},
with proposed explanations based on data geometry and kernel spectra
\citep{bahri2021explaining},
skill-frequency structure
\citep{michaud2023quantization},
training dynamics
\citep{bordelon2024dynamical},
statistical and approximation theory
\citep{havrilla2024scaling},
and representational superposition
\citep{liu2025superposition}.
CFS complements these views by tracking how functional substitutability and independent structure change with scale, linking microscopic organization to macroscopic diminishing returns.


\paragraph{Adaptive computation.}
Transformer efficiency has been improved through dynamic head allocation, token/head pruning, conditional blocks, and adaptive depth
\citep{peng2020mixture,lee2022tokenhead,meng2022adavit,wang2024smarttrim,raposo2024mixturedepths}.
Unlike methods based on learned policies, importance, or compute objectives, CFS-guided routing selects components from predicted pairwise functional substitutability under a fixed budget.

\section{Method}

\subsection{Functional Unit Decomposition}

We decompose Transformer computation into independently intervenable units, using attention heads because their projected outputs form parallel additive paths into the residual stream; the formulation extends to other decomposable components.

For the $l$-th Transformer layer with $H$ attention heads, we write
\begin{equation}
\mathrm{MHA}_l(x)=\sum_{i=1}^{H}c_{l,i}(x),
\quad
c_{l,i}(x)=h_{l,i}(x)W_O^{(i)},
\quad
h_{l,i}(x)=A_{l,i}(x)V_{l,i}(x).
\label{eq:head_decomposition}
\end{equation}
where $W_O^{(i)}$ is the output projection for head $i$, and $c_{l,i}(x)$ its contribution to the residual stream.

We introduce a gate vector $\mathbf{g}=(g_1,\ldots,g_H)$ and write the intervened state as
\begin{equation}
z_l(x;\mathbf{g})
=
r_l(x)+\sum_{i=1}^{H}g_i c_{l,i}(x),
\label{eq:gated_state}
\end{equation}
where $r_l(x)$ contains all residual contributions except the target head outputs; $\mathbf{g}=\mathbf{1}$ recovers the original model, while changing selected gates intervenes on the corresponding units.


We ask not whether states are representationally similar at layer $l$, but whether they remain functionally interchangeable downstream.

\subsection{Conditional Functional Substitutability}

\emph{Conditional Functional Substitutability} (CFS) measures how well one functional unit substitutes for another under a given input.
Let
\begin{equation}
y(x)=F_l\!\left(z_l(x;\mathbf{1})\right),
\qquad
D_i^{\mathrm{drop}}(x)=D\!\left(y(x),y^{-i}(x)\right).
\label{eq:drop_damage}
\end{equation}
where $F_l$ is the downstream network, $y^{-i}(x)$ the output with $g_i=0$, and $D(\cdot,\cdot)$ the output discrepancy. Thus, $D_i^{\mathrm{drop}}(x)$ measures the effect of removing source $i$.

To test whether $j$ can compensate for $i$, we set $g_i=0$, $g_j=\alpha$, keep other gates unchanged, and define
\begin{equation}
\begin{aligned}
D_{i\rightarrow j}(x;\alpha)
&= D\!\left(y(x),y^{i\leftarrow j,\alpha}(x)\right),
\qquad
D^{*}_{i\rightarrow j}(x)
= \min_{\alpha} D_{i\rightarrow j}(x;\alpha), \\
S_{i\rightarrow j}(x)
&= 1-
\frac{D^{*}_{i\rightarrow j}(x)}
     {D_i^{\mathrm{drop}}(x)}.
\end{aligned}
\label{eq:cfs}
\end{equation}

$S_{i\rightarrow j}(x)$ is the fraction of deletion-induced functional damage recovered by $j$: values near $1$ indicate near-complete recovery, while values near $0$ indicate little improvement over removing $i$.


To avoid unstable normalization for negligible deletion effects, we compute CFS only when $D_i^{\mathrm{drop}}(x)>\epsilon$, with $\epsilon$ specified in the experimental protocol.

CFS is \emph{functional}, evaluating downstream behavior rather than representation proximity; \emph{input-conditioned}, so $S_{i\rightarrow j}(x_1)\neq S_{i\rightarrow j}(x_2)$ in general; and \emph{directional}, so $S_{i\rightarrow j}(x)\neq S_{j\rightarrow i}(x)$ in general. It therefore defines a dynamic relation rather than a static component score.

\subsection{Quantifying Functional Organization}

Pairwise CFS induces an input-conditioned functional graph
\begin{equation}
G(x)=\bigl(V,S(x)\bigr),
\qquad
V=\{1,\ldots,H\},
\qquad
S(x)=\left[S_{i\rightarrow j}(x)\right]_{i,j=1}^{H},
\label{eq:cfs_graph}
\end{equation}
where $i\rightarrow j$ indicates that $j$ can substitute for source $i$. We summarize this graph with complementary measures of functional organization.

\paragraph{Raw Oracle.}
For each source component, we measure the strongest available substitution path:
\begin{equation}
O_i^{\mathrm{raw}}(x)=\max_{j\neq i}S_{i\rightarrow j}(x),
\qquad
O^{\mathrm{raw}}=\mathbb{E}_{x,i}\!\left[O_i^{\mathrm{raw}}(x)\right].
\label{eq:raw_oracle}
\end{equation}

Raw Oracle measures the best available substitute. Because additional paths are themselves part of a larger model’s functional organization, it is our primary measure of available substitutability.

\paragraph{Matched Oracle.}
To control candidate count, we restrict each source to a candidate set $\mathcal{C}_m(i)$ of fixed size $m$:
\begin{equation}
O_i^{\mathrm{matched}}(x;m)
=\max_{j\in\mathcal{C}_m(i)}S_{i\rightarrow j}(x),
\quad
O^{\mathrm{matched}}(m)
=\mathbb{E}_{x,i,\mathcal{C}_m}
\!\left[O_i^{\mathrm{matched}}(x;m)\right].
\label{eq:matched_oracle}
\end{equation}
Matched Oracle is therefore a candidate-count control rather than a replacement for Raw Oracle.

\paragraph{Functional Coverage.}
To characterize global compactness beyond pairwise oracle scores, we ask how small a representative set can cover all components under CFS. Given threshold $\tau$, $R\subseteq V$ covers $i$ if $i\in R$ or some retained component substitutes for it:
\begin{equation}
\begin{aligned}
R_{\tau}^{*}(x)
&=\arg\min_{R\subseteq V}|R|
\quad
\text{s.t. }\forall i\in V,\;
i\in R\ \text{or}\ 
\max_{j\in R}S_{i\rightarrow j}(x)\geq\tau,\\
C_{\tau}(x)
&=\frac{|R_{\tau}^{*}(x)|}{H}.
\end{aligned}
\label{eq:functional_coverage}
\end{equation}

Smaller $C_{\tau}$ indicates a more compact functional organization.

\paragraph{Effective Functional Rank.}
We additionally report an effective functional rank 
of the CFS spectrum, providing a continuous measure of spectral compactness without a substitution threshold; the exact estimator follows the experimental setup.

Together, these metrics capture best available substitution, fixed-count substitution density, global functional compactness, and spectral compactness.

\section{Functional Substitutability as a Lens for Neural Redundancy}
\label{sec:redundancy}

CFS asks whether a measurable component effect must be carried by that component itself or can be absorbed by alternative pathways. We first distinguish this relation from importance and similarity, then quantify joint functional reduction under complete functional information.

\subsection{Beyond Importance and Similarity}
\label{sec:beyond_importance_similarity}

\paragraph{Beyond importance.}
If functional substitutability merely reflected source importance, substitute identity should matter little. Across 12 images, 12 layers, and 6 source heads per layer, input-conditioned selection raises recovery from \(0.0015\) to \(0.1213\) without adaptive calibration 
(Table~\ref{tab:cfs_proxy_comparison}(a)). Under the same per-input oracle compensation, functional selection reaches $0.3701$, compared with $0.1988$--$0.2592$ for random, L2, fixed, and Taylor representatives.

Calibration is complementary: for a fixed representative, input-conditioned compensation raises recovery from $0.0015$ to $0.2186$. 
Substitutability therefore depends jointly on source, substitute, and input: a component may be locally effective without being independently indispensable.

\begin{table}[t]
\centering
\footnotesize
\setlength{\tabcolsep}{2.3pt}

\begin{minipage}[t]{0.58\linewidth}
\centering
\textbf{(a) Functional recovery decomposition}\\[2pt]
\begin{tabular}{lcc}
\toprule
Policy & Mean $S$ & Mean KL \\
\midrule
Direct drop                         & 0.0000 & 0.01151 \\
Fixed rep. + fixed $\alpha$         & 0.0015 & 0.01075 \\
L2 rep. + fixed $\alpha$            & 0.0035 & 0.01091 \\
Conditional rep. + fixed $\alpha$   & 0.1213 & 0.00950 \\
Random rep. + oracle $\alpha$       & 0.1988 & 0.00715 \\
L2 rep. + oracle $\alpha$           & 0.2027 & 0.00714 \\
Fixed rep. + oracle $\alpha$        & 0.2186 & 0.00663 \\
Taylor rep. + oracle $\alpha$       & 0.2592 & 0.00551 \\
Conditional rep. + oracle $\alpha$  & \textbf{0.3701} & \textbf{0.00375} \\
\bottomrule
\end{tabular}
\end{minipage}
\hfill
\begin{minipage}[t]{0.39\linewidth}
\centering
\textbf{(b) Proxy correlation with CFS}\\[2pt]
\begin{tabular}{lc}
\toprule
Proxy & Spearman \\
\midrule
Taylor importance        & 0.2386 \\
Attention-map similarity & 0.0175 \\
Contribution cosine      & 0.0087 \\
Contribution proximity ($-$L2) & 0.0003 \\
Parameter cosine         & $-0.0050$ \\
Contribution norm        & $-0.0128$ \\
\bottomrule
\end{tabular}
\end{minipage}

\caption{
CFS versus conventional redundancy proxies.
(a) Functional recovery under representative selection and compensation; conditional policies use per-input oracle information, with the same oracle-$\alpha$ permission for Taylor.
(b) Mean within-source Spearman correlation with CFS over 861 valid ranking units.
}
\label{tab:cfs_proxy_comparison}
\end{table}



\paragraph{Beyond similarity.}


To test whether good substitutes are simply close in parameter or representation space, we rank five candidates for each sample--layer--source unit using conventional proxies and compare them with CFS (Table~\ref{tab:cfs_proxy_comparison}(b)). Parameter-, contribution-, and attention-based similarities are essentially uncorrelated with CFS; even Taylor importance reaches only $0.2386$ Spearman. Current-layer proximity therefore does not reliably predict downstream functional substitutability: nearby states may diverge downstream, while distant states may induce similar behavior.

\paragraph{Input-conditioned and directional structure.}


CFS also varies strongly across inputs. For each fixed $(\text{layer},\text{source})$ pair, the modal substitute covers only $31.25\%$ of 32 inputs, while each source encounters on average $4.97$ distinct best substitutes out of five. 
This variation is structured: agreement falls from $83.3\%$ under mild color perturbations to $50.0\%$ under horizontal flips and $16.7\%$ across images, with corresponding matrix Spearman correlations of $0.955/0.870/0.334$.



CFS is also directional: across 5,760 unordered head pairs, the median absolute gap is $0.0785$ (75th percentile $0.1767$). 
Strongly one-sided relations are uncommon, with $3.94\%$ exceeding $S>0.10$ in one direction and $S\leq0$ in the reverse, and $1.62\%$ exceeding $S>0.25$. 
Thus, CFS is not uniformly highly asymmetric, but cannot generally be represented as an undirected similarity relation.

\subsection{Oracle Functional Reduction Potential}
\label{sec:oracle_reduction}

Pairwise substitutions need not compose under simultaneous intervention because removed components may share substitutes and interact through the nonlinear downstream network. We therefore construct an \emph{Exact Joint-Subset Oracle} that evaluates every retained subset at a fixed budget.

On the final 12-head layer of ViT-B/16, we exhaustively evaluate all subsets retaining 3, 6, or 9 heads ($220/924/220$ subsets) on 448 frozen images. Heads are jointly hard-gated; the oracle minimizes KL to the dense model, while Taylor importance~\citep{michel2019sixteen} uses the same budgets and protocol.

\begin{table}[t]
\centering
\small
\setlength{\tabcolsep}{3pt}
\vspace{-8pt}
\caption{
Exact joint-subset oracle versus Taylor on ViT-B/16.
Fidelity is agreement with the dense prediction; dense accuracy is $81.92\%$.
}
\label{tab:joint_oracle}
\begin{tabular}{c|ccc|ccc}
\toprule
& \multicolumn{3}{c|}{Exact Oracle}
& \multicolumn{3}{c}{Taylor} \\
Keep & KL $\downarrow$ & Acc. & Fid. & KL $\downarrow$ & Acc. & Fid. \\
\midrule
3/12 & \textbf{0.08119} & 81.47 & 96.88 & 0.21156 & 80.58 & 93.75 \\
6/12 & \textbf{0.01290} & 82.37 & 99.55 & 0.05125 & 81.47 & 95.76 \\
9/12 & \textbf{0.00258} & 81.70 & 99.55 & 0.01375 & 81.03 & 98.44 \\
\bottomrule
\end{tabular}
\vspace{-8pt}
\end{table}



The oracle reduces KL by $61.6$--$81.3\%$ relative to Taylor while improving dense-prediction fidelity at every budget (Table~\ref{tab:joint_oracle}); all paired bootstrap 
CIs for oracle-minus-Taylor KL are below zero.



These results expose a gap between \emph{importance redundancy} and \emph{functional redundancy}: importance-based pruning targets components with small individual effects, whereas the oracle selects subsets that collectively preserve full-model behavior. The oracle is an upper bound within the evaluated intervention space rather than a deployable pruning algorithm, motivating Section~\ref{sec:adaptive}: can predicted CFS recover part of this advantage without intervention-time information?

\section{Functional Organization under Neural Scaling}
\label{sec:scaling}

Scaling laws characterize performance growth but not how added computation is organized. We use CFS to study this organization across natural model families, controlled scaling, fixed-capacity architectures, and training dynamics.

\begin{table}[t]
\centering
\small
\setlength{\tabcolsep}{3.2pt}
\caption{
Functional organization across natural scaling families.
Metric is accuracy for ViT and LM loss for BERT/Qwen2.5; Cover/H and Rank/H are normalized.
Qwen2.5-0.5B is omitted due to its 64-d rather than 128-d query heads.
}
\label{tab:natural_scaling}
\begin{tabular}{llrrrrrr}
\toprule
Family & Scale & Params & Metric & Raw & Match-2 & Cover/H & Rank/H \\
\midrule
ViT
& Tiny  & 5.72M   & 76.95 & 0.3379 & 0.3379 & 0.7982 & 0.8998 \\
& Small & 22.05M  & 83.40 & 0.4088 & 0.2852 & 0.7321 & 0.8602 \\
& Base  & 86.57M  & 85.35 & 0.4976 & 0.2889 & 0.5972 & 0.7276 \\
& Large & 304.33M & 86.91 & 0.6034 & 0.3737 & 0.3455 & 0.4647 \\
\midrule
BERT
& Tiny  & 4.42M   & 4.594 & 0.0128 & 0.0128 & 1.0000 & 0.9998 \\
& Small & 28.80M  & 3.033 & 0.1318 & 0.0634 & 0.9922 & 0.9827 \\
& Base  & 109.51M & 3.094 & 0.1365 & 0.0515 & 0.9948 & 0.9769 \\
& Large & 335.17M & 3.102 & 0.1799 & 0.0692 & 0.9414 & 0.9209 \\
\midrule
Qwen2.5
& 1.5B & 1.54B  & 2.702 & 0.2725 & 0.1620 & 0.9549 & 0.9052 \\
& 3B   & 3.09B  & 2.582 & 0.3198 & 0.1953 & 0.8984 & 0.8266 \\
& 7B   & 7.62B  & 2.457 & 0.3712 & 0.2180 & 0.8240 & 0.7427 \\
& 14B  & 14.77B & 2.313 & 0.4430 & 0.2745 & 0.6992 & 0.6419 \\
\bottomrule
\end{tabular}
\vspace{-10pt}
\end{table}

\subsection{Functional Organization across Model Scale}
\label{sec:natural_scaling}

We analyze ViT Tiny--Large on ImageNet-1K
\citep{dosovitskiy2021image,deng2009imagenet},
pretrained BERT Tiny--Large
\citep{devlin2019bert,turc2019wellread},
and Qwen2.5 1.5B--14B
\citep{qwen2024technical}.
Because their tasks and output spaces differ, we compare Raw Oracle, Matched-2 Oracle, normalized functional coverage, and effective functional rank rather than absolute KL.


Table~\ref{tab:natural_scaling} shows a common pattern: Raw Oracle rises while Cover/H and Rank/H fall for ViT and same-granularity Qwen2.5, with a weaker trend in BERT. We exclude Qwen2.5-0.5B because its 64-d query heads differ from the 128-d heads of larger models.

Raw Oracle reflects both substitute quality and candidate availability; we treat the latter as part of scaling because added paths are functional options of the larger model. Matched-2 controls candidate count and is not universally monotonic, showing that scaling changes global organization rather than uniformly strengthening pairwise relations.


Across families, scaling adds functional alternatives without proportional growth in independently indispensable structure: diversity and reuse grow together, providing a microscopic account of diminishing returns without implying that substitutability is their sole origin.

\subsection{Controlled Capacity Scaling}
\label{sec:controlled_scaling}


To isolate scale from pretrained-family differences, we train GPT-2-style models \citep{radford2019language} from scratch with a shared tokenizer, corpus, context length, token budget, and optimization. Width scaling uses 12 layers with 64-d heads and widths $576/768/1152$; depth scaling uses width 768, 12 heads, and $9/12/18$ layers; W768 and D12 are identical.

\begin{table}[t]
\centering
\small
\setlength{\tabcolsep}{3.0pt}
\caption{
Controlled width/depth scaling under a shared training protocol.
W768 and D12 are the same model.
}
\label{tab:controlled_scaling}
\begin{tabular}{llrrrrrr}
\toprule
Axis & Setting & Params & Val. Loss & Raw & Match-2 & Cover/H & Rank/H \\
\midrule
Width
& W576  & 77.40M  & 3.7021 & 0.1882 & 0.1140 & 0.9826 & 0.9550 \\
& W768  & 124.44M & 3.5752 & 0.2125 & 0.1241 & 0.9766 & 0.9335 \\
& W1152 & 250.36M & 3.4000 & 0.2522 & 0.1325 & 0.9497 & 0.8994 \\
\midrule
Depth
& D9  & 103.18M & 3.6252 & 0.2406 & 0.1264 & 0.9236 & 0.9159 \\
& D12 & 124.44M & 3.5752 & 0.2125 & 0.1241 & 0.9766 & 0.9335 \\
& D18 & 166.97M & 3.5024 & 0.2154 & 0.1334 & 0.9722 & 0.9286 \\
\bottomrule
\end{tabular}
\vspace{-8pt}
\end{table}


Table~\ref{tab:controlled_scaling} shows that width scaling follows the natural-family pattern: lower validation loss accompanies higher Raw Oracle and lower Cover/H and Rank/H. Depth scaling differs: validation loss continues to improve while Raw Oracle changes little, from $0.2125$ at D12 to $0.2154$ at D18.

The shared W768/D12 baseline makes this contrast explicit. Widening to W1152 raises Raw Oracle by $0.0397$, whereas deepening to D18 raises it by only $0.0029$; per added parameter, D18 also yields a larger reduction in validation loss. Thus, the scaling path with less growth in functional substitutability is more parameter-efficient, consistent with a larger fraction of added capacity becoming independent functional structure rather than alternative realizations of existing functions.

\subsection{Functional Organization at Fixed Capacity}
\label{sec:fixed_capacity}

We isolate functional organization at fixed nominal capacity using three 12-layer, 768-dimensional GPT-2 models with $6\times128$, $12\times64$, or $24\times32$ attention heads. Parameter counts and dense computation are identical; only head decomposition changes.

\begin{table}[t]
\centering
\small
\setlength{\tabcolsep}{3.2pt}
\caption{
Functional organization at fixed model capacity.
All models have 12 layers, width 768, and identical parameter counts; only head decomposition changes.
}
\label{tab:fixed_capacity}
\begin{tabular}{lrrrrrrr}
\toprule
Heads & Params & Val. Loss & Raw & Match-2 & Cover/H & Rank/H & Eff. Rank \\
\midrule
$6\times128$  & 124.44M & 3.5707 & 0.2012 & 0.1413 & 0.9497 & 0.9507 & 5.704  \\
$12\times64$  & 124.44M & 3.5752 & 0.2125 & 0.1241 & 0.9766 & 0.9335 & 11.202 \\
$24\times32$  & 124.44M & 3.5780 & 0.2583 & 0.1371 & 0.9518 & 0.8725 & 20.939 \\
\bottomrule
\end{tabular}
\vspace{-8pt}
\end{table}

Table~\ref{tab:fixed_capacity} shows that, at identical parameter budgets, validation loss rises from \(3.5707\) to \(3.5780\) as Raw Oracle increases from \(0.2012\) to \(0.2583\) and Rank/H falls from \(0.9507\) to \(0.8725\).
Thus, lower substitutability and higher normalized functional rank are associated with better performance at fixed capacity.
Because head granularity also changes architectural inductive bias, this is associative rather than causal, but shows that parameter count alone does not determine functional organization.

\subsection{Formation and Stabilization during Training}
\label{sec:training_dynamics}

We track CFS at layers 3/7/11 over 100 epochs of ImageNet-100 training for ViT-S AugReg \citep{steiner2021train}, comparing checkpoints with epoch-100 structure on a fixed 512-image probe set.

\begin{figure}[t]
    \centering
    \begin{minipage}[t]{0.54\linewidth}
        \centering
        \includegraphics[width=\linewidth]{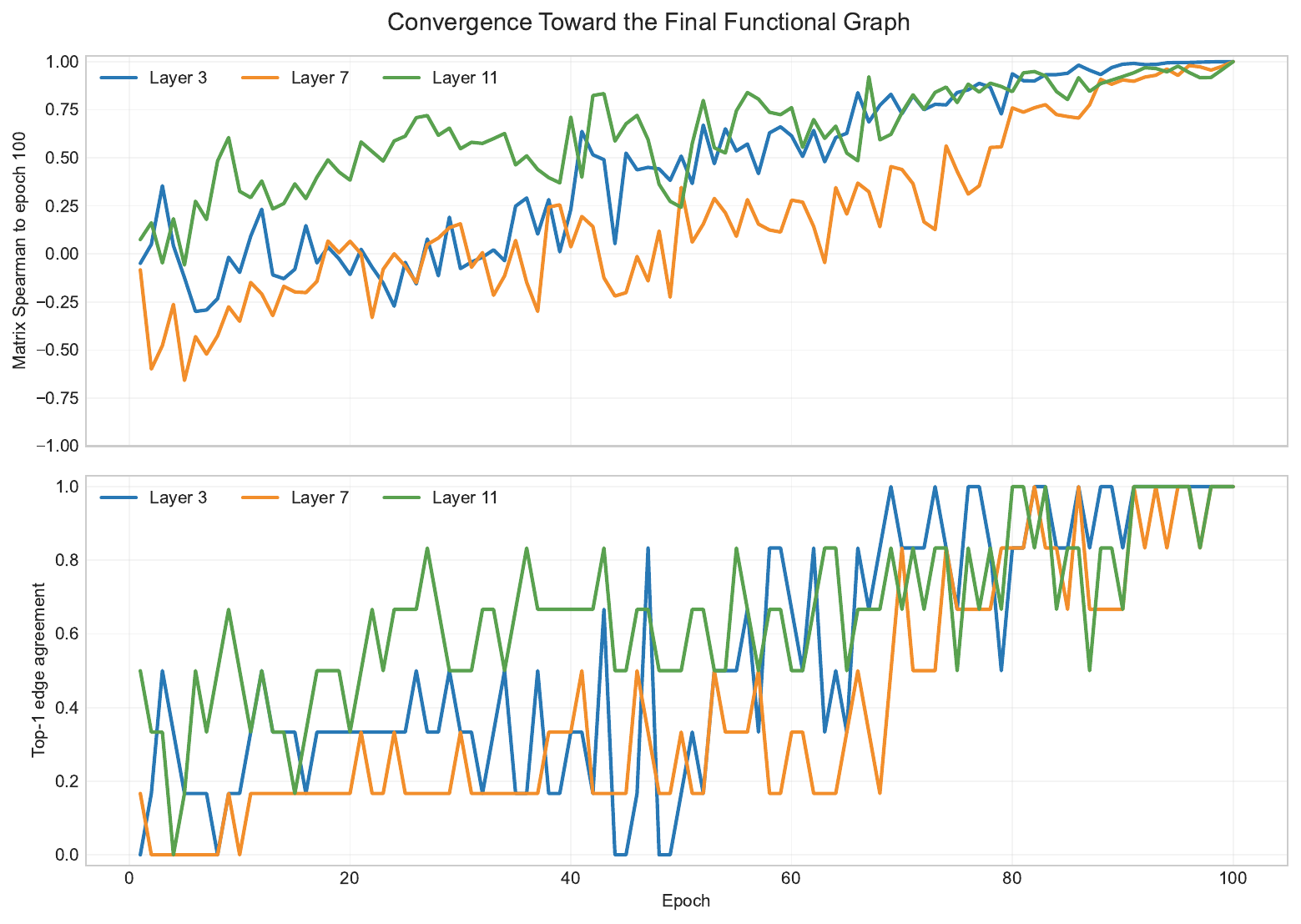}
    \end{minipage}
    \hfill
    \begin{minipage}[t]{0.28\linewidth}
        \centering
        \includegraphics[width=\linewidth]{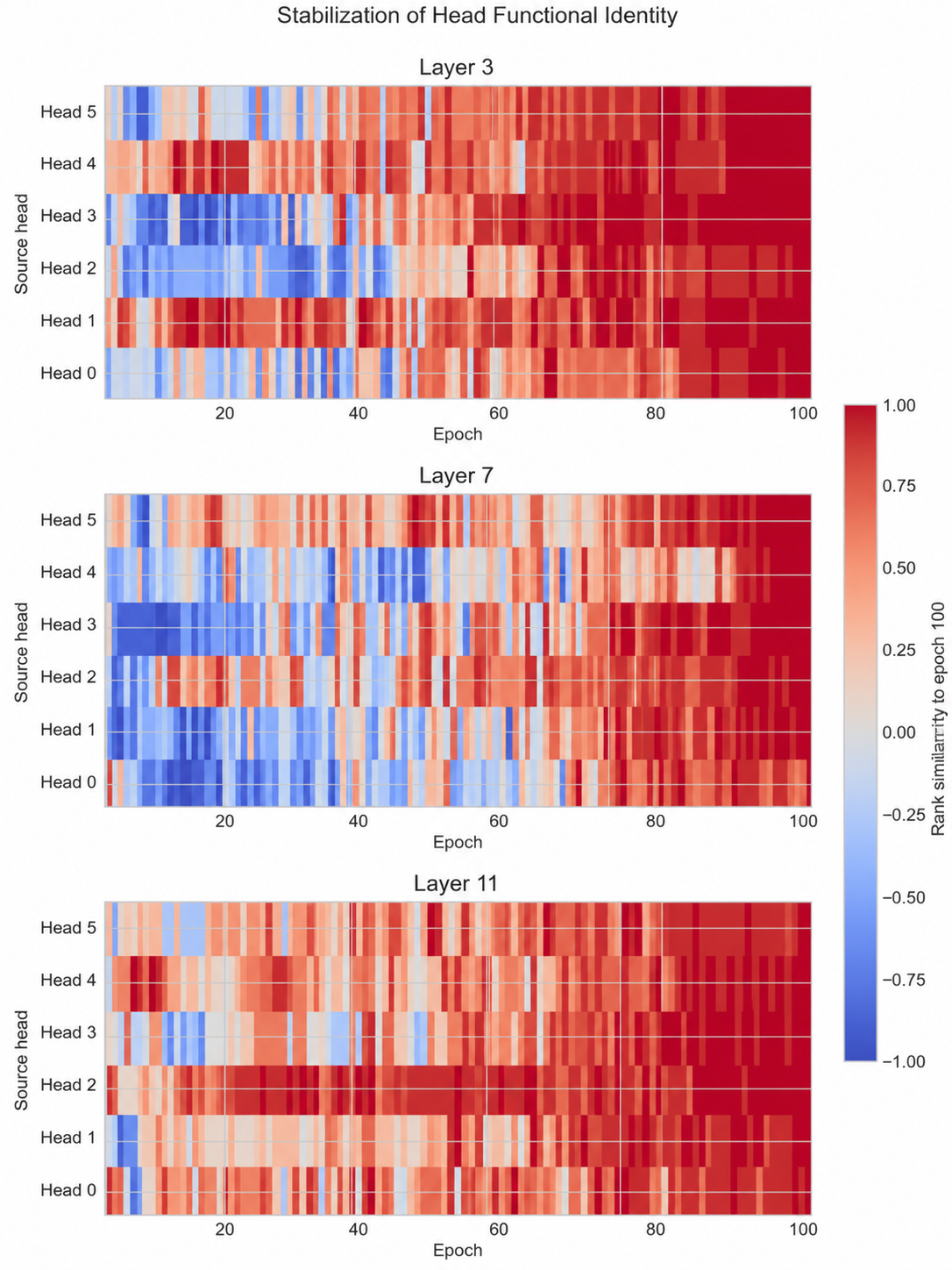}
    \end{minipage}
    \vspace{-6pt}
    \caption{
    \textbf{Formation of functional organization during training.}
    Left: CFS-graph similarity (top) and top-1 substitute agreement (bottom) with epoch 100.
    Right: per-head similarity to final substitute rankings for layers 3, 7, and 11.
    }
    \label{fig:functional_evolution}
    \vspace{-12pt}
\end{figure}


Figure~2 shows gradual but non-monotonic consolidation: layer-11 similarity is \(0.612/0.242/0.787\) at epochs \(25/50/75\), while epoch-75 similarities across layers \(3/7/11\) are \(0.839/0.429/0.787\), revealing heterogeneous stabilization.
Early low CFS does not imply stronger independence, since epoch-1 deletion effects are only $O(10^{-5})$;
optimization progressively shapes both component functions and relations.

\section{From Functional Analysis to Adaptive Computation}
\label{sec:adaptive}

Because CFS requires explicit interventions, we ask whether it can be predicted from forward states and used for input-conditioned allocation without intervention-time information.

\subsection{Observability of Functional Substitutability}
\label{sec:observability}

On frozen DeiT-S \citep{touvron2021training}, we probe CFS from residual states, Q/K statistics, attention logits, value content, aggregated head outputs, and post-$W_O$ contributions, with downstream gradient sensitivity as an analysis-only upper probe. All predictors share targets and three seeds.

\begin{table}[t]
\centering
\small
\setlength{\tabcolsep}{4.2pt}
\caption{
CFS observability on DeiT-S. Recovery is achieved by the predicted representative.
}
\label{tab:observability}
\begin{tabular}{lrrr}
\toprule
Information & Recovery & Rep. Acc. & Spearman \\
\midrule
Residual              & 0.2204 & 36.85 & 0.2553 \\
Q/K moments           & 0.2251 & 38.82 & 0.2718 \\
Attention logits      & 0.2232 & 38.24 & 0.2678 \\
$V$ content           & 0.2245 & 38.65 & 0.2790 \\
$AV$                   & 0.2236 & 38.53 & 0.2732 \\
Post-$W_O$            & 0.2247 & 39.03 & 0.2793 \\
Gradient sensitivity  & \textbf{0.2643} & \textbf{50.68} & \textbf{0.4701} \\
\bottomrule
\end{tabular}
\vspace{-10pt}
\end{table}



Forward features yield similar recovery ($0.220$--$0.225$), whereas downstream sensitivity reaches $0.2643$ recovery and $0.4701$ Spearman; Taylor and the functional oracle achieve $0.2318$ and $0.3231$ recovery, respectively. This gap is consistent with CFS's relational nature: substitutability depends on downstream propagation rather than local geometry alone. The same forward-to-sensitivity gap appears on ViT-S AugReg \citep{steiner2021train}, DINO ViT-S \citep{caron2021emerging}, and BERT on SST-2 and MNLI \citep{devlin2019bert,wang2019glue}, with recovery gains of $0.039$--$0.120$. Gradients remain analysis-only; forward states still provide the routing signal.

\subsection{CFS-Guided Dynamic Functional Routing}
\label{sec:cfs_routing}

We evaluate routing on the final 12-head layer of frozen ViT-B AugReg \citep{steiner2021train} at $K\in\{3,6,9\}$, using 4,096/64/448 train/validation/test images and seeds $17/23/41$. Dense accuracy is $81.92\%$.

\paragraph{Forward-only CFS prediction.}
The router receives only the pre-attention residual tensor
$X_l\in\mathbb{R}^{N\times d}$, summarized as

\begin{equation}
z_l(x)=
\left[
X_{l,\mathrm{CLS}};
\operatorname{Mean}(X_{l,\mathrm{patch}});
\operatorname{Var}(X_{l,\mathrm{patch}})
\right].
\end{equation}

No Q/K/V, head outputs, downstream activations, gradients, labels, or oracle CFS are available. A lightweight encoder $E$ combines this summary with learned head embeddings $e_i,e_j$:

\begin{equation}
\phi_{ij}(x)=
\left[
E(z_l(x));
e_i;
e_j;
|e_i-e_j|;
e_i\odot e_j
\right],
\end{equation}

From $\phi_{ij}(x)$, the router predicts substitutability $\hat S_{i\rightarrow j}(x)$, compensation $\hat\alpha_{i\rightarrow j}(x)$, and source weight $\hat w_i(x)$. The frozen-backbone router has approximately $0.825$M trainable parameters.

\paragraph{Structured subset selection.}
Let $\tilde w_i=\operatorname{softmax}_i(\hat w)$. For retained set $R$, its predicted utility is

\begin{equation}
\hat U(R\mid x)=
\sum_i \tilde w_i(x)
\begin{cases}
1, & i\in R,\\[2pt]
\displaystyle
\max_{j\in R}
\operatorname{clip}
\bigl(\hat S_{i\rightarrow j}(x),0,1\bigr),
& i\notin R .
\end{cases}
\end{equation}




We select $R^*(x)=\arg\max_{|R|=K}\hat U(R\mid x)$ by exact enumeration of all $220/924/220$ subsets for $K=3/6/9$; no greedy approximation is used.

\paragraph{Joint-subset supervision.}
Because pairwise substitutions need not compose, R2 augments R1 with direct joint-subset targets. For each image and budget, a teacher simultaneously gates 48 CFS/router, Taylor/magnitude, similarity, hard-negative, and random masks and records their dense-model KL; R2 aligns predicted subset utility with these outcomes.

\begin{table}[t]
\centering
\small
\setlength{\tabcolsep}{4.5pt}
\caption{
CFS routing versus Taylor under equal logical head budgets. KL is measured against the dense model.
}
\label{tab:routing_main}
\begin{tabular}{rrrrrr}
\toprule
Keep & CFS Acc. & Taylor Acc. & CFS KL & Taylor KL & KL Red. \\
\midrule
3/12 & 81.03 & 80.58 & \textbf{0.1190} & 0.2105 & \textbf{43.46\%} \\
6/12 & 82.59 & 81.47 & \textbf{0.0281} & 0.0513 & \textbf{45.21\%} \\
9/12 & 82.07 & 81.03 & \textbf{0.0070} & 0.0138 & \textbf{49.23\%} \\
\bottomrule
\end{tabular}
\vspace{-12pt}
\end{table}



CFS reduces KL by $43.5$--$49.2\%$ relative to Taylor across all three budgets (Table~\ref{tab:routing_main}), with all paired 95\% confidence intervals below zero. Accuracy trends higher but is not statistically resolved, so output fidelity is the primary evidence. Under shared learned calibration, CFS retains KL advantages of $0.0696/0.0178/0.0068$ at $K=3/6/9$; with oracle pairwise $\alpha$, they remain $0.0604/0.0154/0.0049$, with all six confidence intervals excluding zero. The gain therefore comes from functional subset selection rather than calibration alone.



R2 further reduces KL over R1 from $0.1217/0.0315/0.00874$ to $0.1190/0.0281/0.00699$ at $K=3/6/9$, with resolved gains at $K=6,9$ and a borderline gain at $K=3$, confirming higher-order interactions beyond pairwise CFS. Online and frozen offline masks agree on $99.78\%$ of images at every budget. Because the current backend still executes dense fused QKV and output-projection kernels, these are logical rather than wall-clock budgets; our claim is better preservation of dense-model behavior under equal component budgets.

\section{Discussion}
\label{sec:discussion}

We treat attention heads as functional units; extending CFS to MLP blocks, experts, tokens, and cross-layer paths could test its generality across granularities. 
The gap between pairwise CFS and joint interventions further motivates explicit modeling of higher-order functional interactions.

Our results suggest functional organization as a design variable beyond parameter count, but the fixed-capacity study changes architectural decomposition and does not establish causal benefits from optimizing CFS structure.
A natural next step is to regularize functional organization while holding architecture and parameter count fixed, testing whether more independent functional structure can improve effective capacity per parameter and ultimately alter the empirical scaling curve. On the systems side, routing improves preservation under logical budgets rather than wall-clock speed; computational gains require backends that exploit input-dependent structured sparsity.


\section{Conclusion}
\label{sec:conclusion}

We introduced Conditional Functional Substitutability (CFS), which characterizes input-conditioned, directional functional redundancy by asking whether one component can reproduce another's downstream role. Across Transformer vision and language models, CFS reveals systematic functional reorganization with scale; controlled experiments further link more independent functional structure to greater scaling efficiency and better performance at fixed capacity. Predicted CFS also enables dynamic selection that better preserves dense-model behavior than importance-based routing under equal logical budgets. Together, these results establish functional substitutability as a unified perspective on neural redundancy, scaling, and adaptive computation.

\subsection*{AI use statement}

We used generative AI tools to provide feedback on research methodology and
experimental presentation, assist in interpreting experimental results, support
code implementation and debugging, and assist with manuscript drafting and editing.
We additionally used generative AI for literature search and summarization,
identifying relevant work, and refining figures and paper structure.
All AI-assisted suggestions were independently reviewed by the authors; cited
literature was verified against original sources, AI-assisted code was reviewed
and tested against the intended experimental behavior, reported numerical results
were checked against our experimental outputs, and all methodological decisions
and final claims were made by the authors.
We take responsibility for the final content of this work, including text, claims,
code, and artifacts produced with the aid of generative AI.

\subsection*{Reproducibility statement}

We provide experimental protocols, metric definitions, implementation details,
and additional analyses in the appendix.
Appendix~A provides additional training-dynamics results and visualization details.
Appendix~B specifies the CFS evaluation protocol, validity criterion, cross-task
functional outcomes, aggregation rules, and the exact definition of effective
functional rank.
Appendix~C details the conventional redundancy proxies and additional analyses of
input conditioning and directionality.
Appendix~D provides the exhaustive joint-subset oracle protocol and statistical
tests.
Appendix~E describes the natural-family comparisons, controlled width/depth scaling,
and fixed-capacity head-granularity experiments.
Appendix~F provides additional routing supervision, ablations, confidence intervals,
and execution details.
The main text and appendix report the evaluated model families, datasets, intervention
units, candidate-selection protocols, retained-component budgets, evaluation splits,
and random seeds where applicable.
Code, evaluation scripts, and experimental configurations will be released upon
acceptance.



\bibliography{iclr2027_conference}

@inproceedings{michel2019sixteen,
  author    = {Michel, Paul and Levy, Omer and Neubig, Graham},
  title     = {Are Sixteen Heads Really Better than One?},
  booktitle = {Advances in Neural Information Processing Systems 32},
  pages     = {14014--14024},
  year      = {2019},
  url       = {https://dblp.org/rec/conf/nips/MichelLN19}
}

@inproceedings{voita2019analyzing,
  author    = {Voita, Elena and Talbot, David and Moiseev, Fedor and Sennrich, Rico and Titov, Ivan},
  title     = {Analyzing Multi-Head Self-Attention: Specialized Heads Do the Heavy Lifting, the Rest Can Be Pruned},
  booktitle = {Proceedings of the 57th Annual Meeting of the Association for Computational Linguistics},
  pages     = {5797--5808},
  year      = {2019},
  url       = {https://dblp.org/rec/conf/acl/VoitaTMST19}
}

@inproceedings{budhraja2020weak,
  author    = {Budhraja, Aakriti and Pande, Madhura and Nema, Preksha and Kumar, Pratyush and Khapra, Mitesh M.},
  title     = {On the Weak Link between Importance and Prunability of Attention Heads},
  booktitle = {Proceedings of the 2020 Conference on Empirical Methods in Natural Language Processing},
  pages     = {3230--3235},
  year      = {2020},
  url       = {https://dblp.org/rec/conf/emnlp/BudhrajaPNKK20}
}

@inproceedings{ma2023llmpruner,
  author    = {Ma, Xinyin and Fang, Gongfan and Wang, Xinchao},
  title     = {{LLM-Pruner}: On the Structural Pruning of Large Language Models},
  booktitle = {Advances in Neural Information Processing Systems 36},
  year      = {2023},
  url       = {https://dblp.org/rec/conf/nips/MaFW23}
}

@inproceedings{men2024shortgpt,
  author    = {Men, Xin and Xu, Mingyu and Zhang, Qingyu and Yuan, Qianhao and Wang, Bingning and Lin, Hongyu and Lu, Yaojie and Han, Xianpei and Chen, Weipeng},
  title     = {{ShortGPT}: Layers in Large Language Models are More Redundant Than You Expect},
  booktitle = {Findings of the Association for Computational Linguistics: ACL 2025},
  pages     = {20192--20204},
  year      = {2025},
  doi       = {10.18653/v1/2025.findings-acl.1035},
  url       = {https://dblp.org/rec/conf/acl/MenXZYWL0HC25}
}

@inproceedings{dalvi2020redundancy,
  author    = {Dalvi, Fahim and Sajjad, Hassan and Durrani, Nadir and Belinkov, Yonatan},
  title     = {Analyzing Redundancy in Pretrained Transformer Models},
  booktitle = {Proceedings of the 2020 Conference on Empirical Methods in Natural Language Processing},
  pages     = {4908--4926},
  year      = {2020},
  url       = {https://dblp.org/rec/conf/emnlp/DalviSDB20}
}

@inproceedings{bian2021attention,
  author    = {Bian, Yuchen and Huang, Jiaji and Cai, Xingyu and Yuan, Jiahong and Church, Kenneth},
  title     = {On Attention Redundancy: A Comprehensive Study},
  booktitle = {Proceedings of the 2021 Conference of the North American Chapter of the Association for Computational Linguistics: Human Language Technologies},
  pages     = {930--945},
  year      = {2021},
  url       = {https://dblp.org/rec/conf/naacl/BianHCYC21}
}

@article{bhojanapalli2021reuse,
  author  = {Bhojanapalli, Srinadh and Chakrabarti, Ayan and Veit, Andreas and Lukasik, Michal and Jain, Himanshu and Liu, Frederick and Chang, Yin-Wen and Kumar, Sanjiv},
  title   = {Leveraging Redundancy in Attention with Reuse Transformers},
  journal = {CoRR},
  volume  = {abs/2110.06821},
  year    = {2021},
  url     = {https://dblp.org/rec/journals/corr/abs-2110-06821}
}

@inproceedings{bansal2021revisiting,
  author    = {Bansal, Yamini and Nakkiran, Preetum and Barak, Boaz},
  title     = {Revisiting Model Stitching to Compare Neural Representations},
  booktitle = {Advances in Neural Information Processing Systems 34},
  pages     = {225--236},
  year      = {2021},
  url       = {https://dblp.org/rec/conf/nips/BansalNB21}
}

@inproceedings{zhang2023patching,
  author    = {Zhang, Fred and Nanda, Neel},
  title     = {Towards Best Practices of Activation Patching in Language Models: Metrics and Methods},
  booktitle = {The Twelfth International Conference on Learning Representations},
  year      = {2024},
  url       = {https://dblp.org/rec/conf/iclr/ZhangN24}
}

@article{mcgrath2023hydra,
  author  = {McGrath, Thomas and Rahtz, Matthew and Kram{\'a}r, J{\'a}nos and Mikulik, Vladimir and Legg, Shane},
  title   = {The Hydra Effect: Emergent Self-Repair in Language Model Computations},
  journal = {CoRR},
  volume  = {abs/2307.15771},
  year    = {2023},
  url     = {https://dblp.org/rec/journals/corr/abs-2307-15771}
}

@inproceedings{rushing2024selfrepair,
  author    = {Rushing, Cody and Nanda, Neel},
  title     = {Explorations of Self-Repair in Language Models},
  booktitle = {Proceedings of the 41st International Conference on Machine Learning},
  pages     = {42836--42855},
  year      = {2024},
  url       = {https://dblp.org/rec/conf/icml/RushingN24}
}

@article{gong2026coax,
  author  = {Gong, Zhiren and Zeng, Zihao and Yuen, Chau and Lim, Wei Yang Bryan},
  title   = {Conditional Co-Ablation: Recovering Self-Repair Backups in Transformer Circuits},
  journal = {CoRR},
  volume  = {abs/2607.01940},
  year    = {2026},
  url     = {https://dblp.org/rec/journals/corr/abs-2607-01940}
}

@article{kaplan2020scaling,
  author  = {Kaplan, Jared and McCandlish, Sam and Henighan, Tom and Brown, Tom B. and Chess, Benjamin and Child, Rewon and Gray, Scott and Radford, Alec and Wu, Jeffrey and Amodei, Dario},
  title   = {Scaling Laws for Neural Language Models},
  journal = {CoRR},
  volume  = {abs/2001.08361},
  year    = {2020},
  url     = {https://dblp.org/rec/journals/corr/abs-2001-08361}
}

@article{hoffmann2022compute,
  author  = {Hoffmann, Jordan and Borgeaud, Sebastian and Mensch, Arthur and Buchatskaya, Elena and Cai, Trevor and Rutherford, Eliza and de Las Casas, Diego and Hendricks, Lisa Anne and Welbl, Johannes and Clark, Aidan and Hennigan, Tom and Noland, Eric and Millican, Katie and van den Driessche, George and Damoc, Bogdan and Guy, Aurelia and Osindero, Simon and Simonyan, Karen and Elsen, Erich and Rae, Jack W. and Vinyals, Oriol and Sifre, Laurent},
  title   = {Training Compute-Optimal Large Language Models},
  journal = {CoRR},
  volume  = {abs/2203.15556},
  year    = {2022},
  url     = {https://dblp.org/rec/journals/corr/abs-2203-15556}
}

@article{bahri2021explaining,
  author  = {Bahri, Yasaman and Dyer, Ethan and Kaplan, Jared and Lee, Jaehoon and Sharma, Utkarsh},
  title   = {Explaining Neural Scaling Laws},
  journal = {CoRR},
  volume  = {abs/2102.06701},
  year    = {2021},
  url     = {https://dblp.org/rec/journals/corr/abs-2102-06701}
}

@inproceedings{michaud2023quantization,
  author    = {Michaud, Eric J. and Liu, Ziming and Girit, Uzay and Tegmark, Max},
  title     = {The Quantization Model of Neural Scaling},
  booktitle = {Advances in Neural Information Processing Systems 36},
  year      = {2023},
  url       = {https://dblp.org/rec/conf/nips/MichaudLGT23}
}

@inproceedings{bordelon2024dynamical,
  author    = {Bordelon, Blake and Atanasov, Alexander B. and Pehlevan, Cengiz},
  title     = {A Dynamical Model of Neural Scaling Laws},
  booktitle = {Proceedings of the 41st International Conference on Machine Learning},
  pages     = {4345--4382},
  year      = {2024},
  url       = {https://dblp.org/rec/conf/icml/BordelonAP24}
}

@inproceedings{havrilla2024scaling,
  author    = {Havrilla, Alexander and Liao, Wenjing},
  title     = {Understanding Scaling Laws with Statistical and Approximation Theory for Transformer Neural Networks on Intrinsically Low-dimensional Data},
  booktitle = {Advances in Neural Information Processing Systems 37},
  year      = {2024},
  url       = {https://dblp.org/rec/conf/nips/HavrillaL24}
}

@inproceedings{liu2025superposition,
  author    = {Liu, Yizhou and Liu, Ziming and Gore, Jeff},
  title     = {Superposition Yields Robust Neural Scaling},
  booktitle = {Advances in Neural Information Processing Systems 38},
  year      = {2025},
  url       = {https://dblp.org/rec/conf/nips/LiuLG25}
}

@inproceedings{peng2020mixture,
  author    = {Peng, Hao and Schwartz, Roy and Li, Dianqi and Smith, Noah A.},
  title     = {A Mixture of h-1 Heads is Better than h Heads},
  booktitle = {Proceedings of the 58th Annual Meeting of the Association for Computational Linguistics},
  pages     = {6566--6577},
  year      = {2020},
  url       = {https://dblp.org/rec/conf/acl/PengSLS20}
}

@inproceedings{lee2022tokenhead,
  author    = {Lee, Chonghan and Khan, Md Fahim Faysal and Brufau, Rita Brugarolas and Ding, Ke and Narayanan, Vijaykrishnan},
  title     = {Token and Head Adaptive Transformers for Efficient Natural Language Processing},
  booktitle = {Proceedings of the 29th International Conference on Computational Linguistics},
  pages     = {4575--4584},
  year      = {2022},
  url       = {https://dblp.org/rec/conf/coling/LeeKBDN22}
}

@inproceedings{meng2022adavit,
  author    = {Meng, Lingchen and Li, Hengduo and Chen, Bor-Chun and Lan, Shiyi and Wu, Zuxuan and Jiang, Yu-Gang and Lim, Ser-Nam},
  title     = {{AdaViT}: Adaptive Vision Transformers for Efficient Image Recognition},
  booktitle = {IEEE/CVF Conference on Computer Vision and Pattern Recognition},
  pages     = {12299--12308},
  year      = {2022},
  doi       = {10.1109/CVPR52688.2022.01199},
  url       = {https://dblp.org/rec/conf/cvpr/MengLCLWJL22}
}

@inproceedings{wang2024smarttrim,
  author    = {Wang, Zekun and Chen, Jingchang and Zhou, Wangchunshu and Zhu, Haichao and Liang, Jiafeng and Shan, Liping and Liu, Ming and Xu, Dongliang and Yang, Qing and Qin, Bing},
  title     = {{SmartTrim}: Adaptive Tokens and Attention Pruning for Efficient Vision-Language Models},
  booktitle = {Proceedings of the 2024 Joint International Conference on Computational Linguistics, Language Resources and Evaluation},
  pages     = {14937--14953},
  year      = {2024},
  url       = {https://dblp.org/rec/conf/coling/WangCZZLS0X0024}
}

@article{raposo2024mixturedepths,
  author  = {Raposo, David and Ritter, Samuel and Richards, Blake A. and Lillicrap, Timothy P. and Humphreys, Peter Conway and Santoro, Adam},
  title   = {Mixture-of-Depths: Dynamically Allocating Compute in Transformer-Based Language Models},
  journal = {CoRR},
  volume  = {abs/2404.02258},
  year    = {2024},
  url     = {https://dblp.org/rec/journals/corr/abs-2404-02258}
}

@inproceedings{dosovitskiy2021image,
  author    = {Dosovitskiy, Alexey and Beyer, Lucas and Kolesnikov, Alexander and Weissenborn, Dirk and Zhai, Xiaohua and Unterthiner, Thomas and Dehghani, Mostafa and Minderer, Matthias and Heigold, Georg and Gelly, Sylvain and Uszkoreit, Jakob and Houlsby, Neil},
  title     = {An Image is Worth 16x16 Words: Transformers for Image Recognition at Scale},
  booktitle = {The Ninth International Conference on Learning Representations},
  year      = {2021},
  url       = {https://dblp.org/rec/conf/iclr/DosovitskiyB0WZ21}
}

@inproceedings{deng2009imagenet,
  author    = {Deng, Jia and Dong, Wei and Socher, Richard and Li, Li-Jia and Li, Kai and Fei-Fei, Li},
  title     = {{ImageNet}: A Large-Scale Hierarchical Image Database},
  booktitle = {IEEE Conference on Computer Vision and Pattern Recognition},
  pages     = {248--255},
  year      = {2009},
  doi       = {10.1109/CVPR.2009.5206848},
  url       = {https://dblp.org/rec/conf/cvpr/DengDSLL009}
}

@inproceedings{devlin2019bert,
  author    = {Devlin, Jacob and Chang, Ming-Wei and Lee, Kenton and Toutanova, Kristina},
  title     = {{BERT}: Pre-training of Deep Bidirectional Transformers for Language Understanding},
  booktitle = {Proceedings of the 2019 Conference of the North American Chapter of the Association for Computational Linguistics: Human Language Technologies},
  pages     = {4171--4186},
  year      = {2019},
  doi       = {10.18653/v1/N19-1423},
  url       = {https://dblp.org/rec/conf/naacl/DevlinCLT19}
}

@article{turc2019wellread,
  author  = {Turc, Iulia and Chang, Ming-Wei and Lee, Kenton and Toutanova, Kristina},
  title   = {Well-Read Students Learn Better: The Impact of Student Initialization on Knowledge Distillation},
  journal = {CoRR},
  volume  = {abs/1908.08962},
  year    = {2019},
  url     = {https://dblp.org/rec/journals/corr/abs-1908-08962}
}

@article{qwen2024technical,
  author  = {Yang, An and Yang, Baosong and Zhang, Beichen and Hui, Binyuan and Zheng, Bo and Yu, Bowen and Li, Chengyuan and Liu, Dayiheng and Huang, Fei and Wei, Haoran and Lin, Huan and Yang, Jian and Tu, Jianhong and Zhang, Jianwei and Yang, Jianxin and Yang, Jiaxi and Zhou, Jingren and Lin, Junyang and Dang, Kai and Lu, Keming and Bao, Keqin and Yang, Kexin and Yu, Le and Li, Mei and Xue, Mingfeng and Zhang, Pei and Zhu, Qin and Men, Rui and Lin, Runji and Li, Tianhao and Xia, Tingyu and Ren, Xingzhang and Ren, Xuancheng and Fan, Yang and Su, Yang and Zhang, Yichang and Wan, Yu and Liu, Yuqiong and Cui, Zeyu and Zhang, Zhenru and Qiu, Zihan},
  title   = {{Qwen2.5} Technical Report},
  journal = {CoRR},
  volume  = {abs/2412.15115},
  year    = {2024},
  url     = {https://dblp.org/rec/journals/corr/abs-2412-15115}
}

@article{radford2019language,
  author = {Radford, Alec and Wu, Jeffrey and Child, Rewon and Luan, David and Amodei, Dario and Sutskever, Ilya},
  title  = {Language Models are Unsupervised Multitask Learners},
  year   = {2019}
}

@article{steiner2021train,
  author  = {Steiner, Andreas and Kolesnikov, Alexander and Zhai, Xiaohua and Wightman, Ross and Uszkoreit, Jakob and Beyer, Lucas},
  title   = {How to Train Your {ViT}? Data, Augmentation, and Regularization in Vision Transformers},
  journal = {CoRR},
  volume  = {abs/2106.10270},
  year    = {2021},
  url     = {https://dblp.org/rec/journals/corr/abs-2106-10270}
}

@inproceedings{touvron2021training,
  author    = {Touvron, Hugo and Cord, Matthieu and Douze, Matthijs and Massa, Francisco and Sablayrolles, Alexandre and J{\'e}gou, Herv{\'e}},
  title     = {Training Data-Efficient Image Transformers {\&} Distillation Through Attention},
  booktitle = {Proceedings of the 38th International Conference on Machine Learning},
  pages     = {10347--10357},
  year      = {2021},
  url       = {https://dblp.org/rec/conf/icml/TouvronCDMSJ21}
}

@inproceedings{caron2021emerging,
  author    = {Caron, Mathilde and Touvron, Hugo and Misra, Ishan and J{\'e}gou, Herv{\'e} and Mairal, Julien and Bojanowski, Piotr and Joulin, Armand},
  title     = {Emerging Properties in Self-Supervised Vision Transformers},
  booktitle = {IEEE/CVF International Conference on Computer Vision},
  pages     = {9630--9640},
  year      = {2021},
  doi       = {10.1109/ICCV48922.2021.00951},
  url       = {https://dblp.org/rec/conf/iccv/CaronTMJMBJ21}
}

@inproceedings{wang2019glue,
  author    = {Wang, Alex and Singh, Amanpreet and Michael, Julian and Hill, Felix and Levy, Omer and Bowman, Samuel R.},
  title     = {{GLUE}: A Multi-Task Benchmark and Analysis Platform for Natural Language Understanding},
  booktitle = {The Seventh International Conference on Learning Representations},
  year      = {2019},
  url       = {https://dblp.org/rec/conf/iclr/WangSMHLB19}
}
\bibliographystyle{iclr2027_conference}

\appendix

\section{Additional Training-Dynamics Analysis}
\label{app:training_dynamics}

We provide additional visualizations complementing the training-dynamics
analysis in Section~\ref{sec:training_dynamics}. These results illustrate the
evolution of predictive performance, functional substitutability, functional
compactness, and the conditional substitution graph throughout optimization.

The trajectory is measured on a single ViT-S AugReg training run over 100
epochs, using the same fixed 512-image probe set at layers 3, 7, and 11.
Epochs 1--3 have deletion effects close to the numerical validity threshold
and should be interpreted as warm-up behavior rather than evidence of a sharp
functional transition. Likewise, similarity to epoch 100 measures convergence
toward the final checkpoint rather than intrinsic model quality.

\begin{figure}[t]
    \centering
    \includegraphics[width=\linewidth]{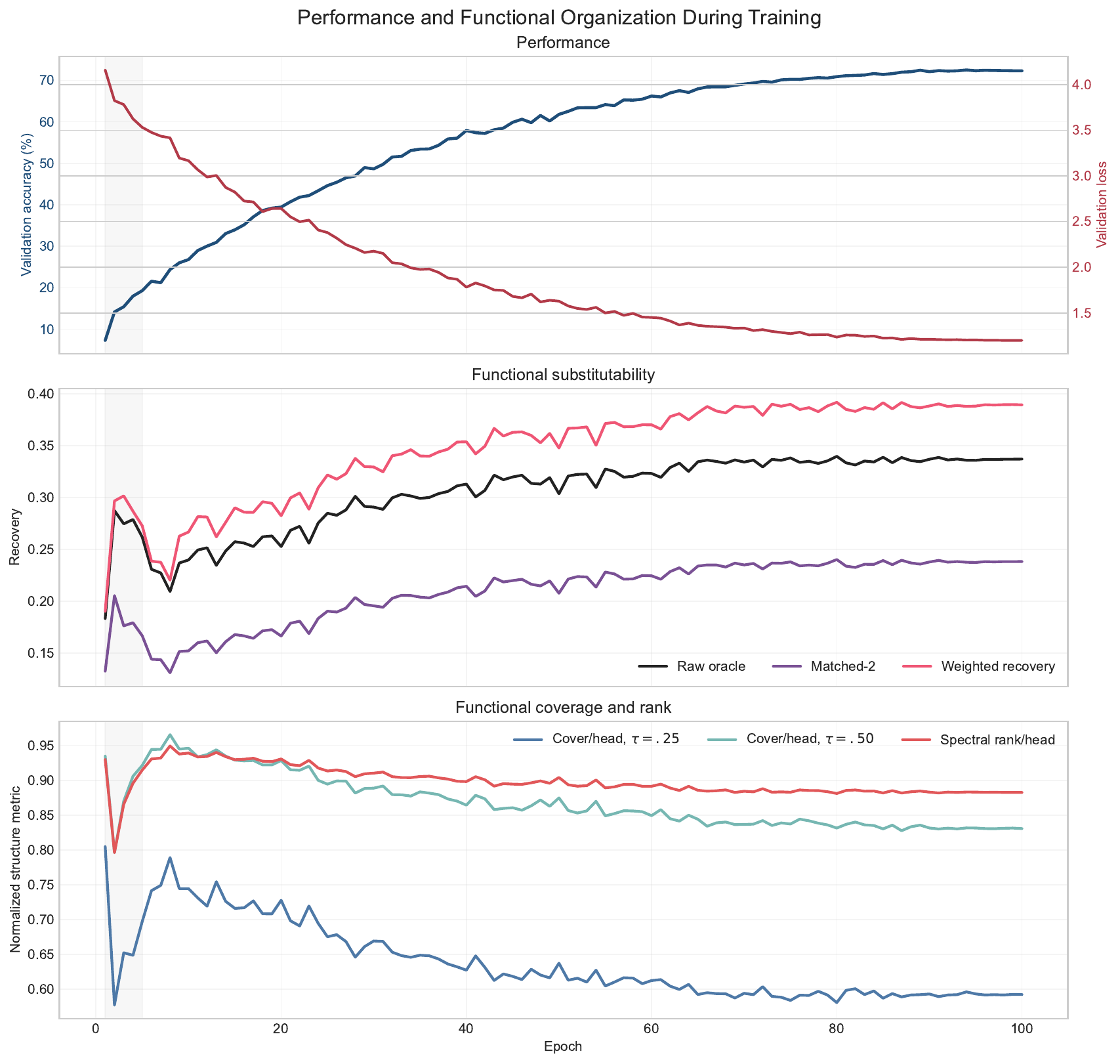}
    \caption{
    \textbf{Performance and functional organization throughout training.}
    Validation performance improves steadily, while functional
    substitutability, coverage, and effective rank follow distinct trajectories.
    Predictive learning and functional reorganization are therefore not
    synchronized.
    }
    \label{fig:app_trajectory_metrics}
\end{figure}

\begin{figure}[t]
    \centering
    \includegraphics[width=\linewidth]{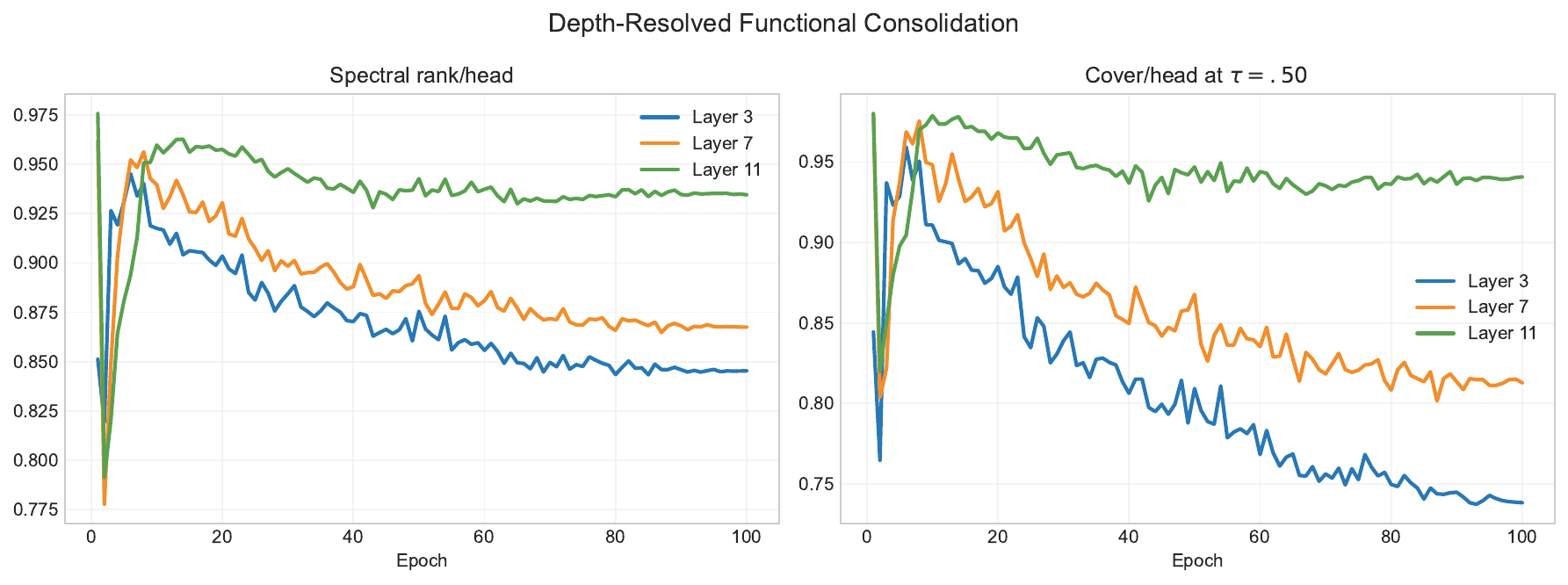}
    \caption{
    \textbf{Depth-resolved functional consolidation.}
    Effective functional rank and functional coverage evolve differently across
    layers 3, 7, and 11, indicating that functional consolidation proceeds at
    different rates and to different degrees throughout the network.
    }
    \label{fig:app_rank_cover}
\end{figure}

\begin{figure}[t]
    \centering
    \includegraphics[width=\linewidth]{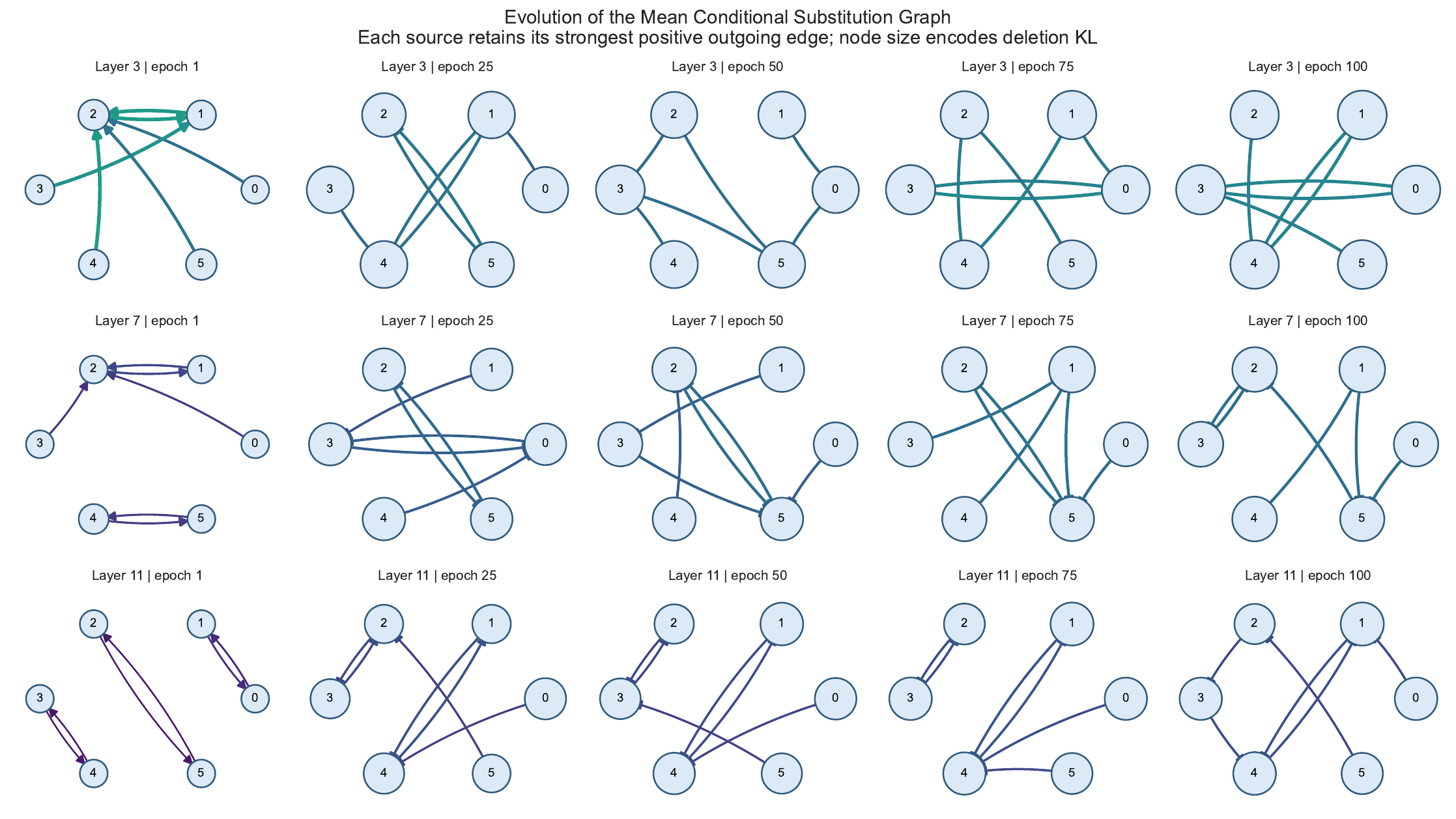}
    \caption{
    \textbf{Evolution of the mean CFS graph during training.}
    We visualize layers 3, 7, and 11 at epochs 1, 25, 50, 75, and 100.
    For each source head, only its strongest positive outgoing substitution edge
    is shown; node size reflects mean deletion sensitivity.
    These snapshots provide a qualitative view of the progressive reorganization
    quantified in Section~\ref{sec:training_dynamics}.
    }
    \label{fig:app_graph_snapshots}
\end{figure}

The graph snapshots are dataset-level summaries of the conditional CFS graph:
the displayed strongest edge need not be selected for every individual input.
Together with Figure~\ref{fig:functional_evolution}, these additional views show
that functional organization changes along multiple axes during training.
Performance can improve while substitution strength, graph compactness, and
component roles remain in motion, and different layers consolidate at
different rates.

\FloatBarrier

\section{Additional CFS Definitions and Protocol Details}
\label{app:cfs_details}

\subsection{Aggregation, Validity, and Functional Outcomes}
\label{app:cfs_protocol}

CFS is computed at the input level before any dataset averaging. In particular,
\begin{equation}
\mathbb{E}_{x}\!\left[
1-\frac{D^{*}_{i\rightarrow j}(x)}
{D_i^{\mathrm{drop}}(x)}
\right]
\neq
1-
\frac{\mathbb{E}_{x}[D^{*}_{i\rightarrow j}(x)]}
{\mathbb{E}_{x}[D_i^{\mathrm{drop}}(x)]},
\end{equation}
and we use the left-hand side throughout. This preserves the conditional nature of the relation rather than first collapsing intervention effects across inputs. Macro statistics across layers are computed as equal-weight means of layer-level summaries rather than by pooling all source units.

Unless otherwise noted, a source--input pair is considered valid only when its single-unit deletion effect satisfies
\begin{equation}
D_i^{\mathrm{drop}}(x)>10^{-5}.
\end{equation}
This filtering avoids unstable normalization when the source has a numerically negligible effect.

The discrepancy $D$ is always measured in the model's native predictive output space. For image classification, we use KL divergence between the dense and intervened class distributions. For BERT, the functional outcome is the mean KL divergence over the masked-token predictive distributions. For causal language models, we use mean KL over selected next-token distributions. Because these output spaces differ, absolute KL values are not compared across model families; CFS recovery, normalized coverage, and normalized effective rank retain common semantics across families.

\begin{table}[t]
\centering
\small
\setlength{\tabcolsep}{3.2pt}
\caption{
CFS protocol for the natural-family and training-dynamics experiments.
Matched-2 denotes the expected best substitute among two candidates.
}
\label{tab:app_cfs_protocol}
\begin{tabular}{llllll}
\toprule
Setting & CFS samples & Evaluated layers & $\alpha$ search & Valid source & Coverage \\
\midrule
ViT natural
& 512
& final block
& 31-point $[0,3]$
& $D^{\mathrm{drop}}>10^{-5}$
& $\tau=0.5$
\\
BERT natural
& 32
& final layer
& 16-point grid
& $D^{\mathrm{drop}}>10^{-5}$
& $\tau=0.5$
\\
Qwen2.5 natural
& 32
& $\approx25/50/75\%$ depth
& 16-point grid
& $D^{\mathrm{drop}}>10^{-5}$
& $\tau=0.5$
\\
Training dynamics
& 512
& layers $3/7/11$
& 16-point $[0,3]$
& $D^{\mathrm{drop}}>10^{-5}$
& $\tau=0.25/0.5/0.75$
\\
\bottomrule
\end{tabular}
\end{table}

For the natural ViT family, CFS is evaluated on a shared aligned image subset and KL is measured over the 1000-way class distribution. BERT uses a frozen MNLI validation-text subset as unlabeled text, with eight deterministically selected tokens masked per text; CFS is computed on a fixed 32-text subset and the reported MLM loss on 512 texts. Qwen2.5 uses a frozen FineWeb-Edu slice with sequence length 128; CFS is evaluated on 32 contexts, while language-model loss is computed separately over the larger frozen evaluation slice.

\subsection{Directed Spectral Effective Rank}
\label{app:effective_rank}

Functional coverage is thresholded, so we additionally use a continuous spectral statistic. For each input, we construct a nonnegative directed affinity matrix
\begin{equation}
F_{ij}(x)=
\begin{cases}
\operatorname{clip}\!\left(S_{i\rightarrow j}(x),0,1\right),
& i\neq j,\; i\text{ valid},\\
1,
& i=j,\; i\text{ valid},\\
0,
& i\text{ invalid}.
\end{cases}
\end{equation}
Let $\sigma_1,\ldots,\sigma_H$ be the singular values of $F(x)$ and
\begin{equation}
p_r=\frac{\sigma_r}{\sum_q\sigma_q}.
\end{equation}
We define
\begin{equation}
R_{\mathrm{eff}}(x)
=
\exp\!\left(
-\sum_{r:p_r>0}p_r\log p_r
\right),
\qquad
\mathrm{Rank/H}
=
\frac{\mathbb{E}_x[R_{\mathrm{eff}}(x)]}{H}.
\end{equation}

Using singular values allows the statistic to retain the directional CFS matrix without symmetrization. Rank/H is a descriptive measure of spectral compactness rather than a claim about the exact intrinsic functional dimension of the network. Lower Rank/H indicates that the directed substitution structure is concentrated into a smaller effective spectral support.

Similarly,
\begin{equation}
\mathrm{Cover/H}
=
\frac{\mathbb{E}_x[|R_\tau^*(x)|]}{H}
\end{equation}
normalizes the exact minimum representative cover by the number of functional units. Lower Cover/H therefore indicates stronger functional compression at the chosen recovery threshold; it does not by itself imply better task performance.

\section{Additional Redundancy Analyses}
\label{app:redundancy_details}

\subsection{Conventional Proxy Definitions}
\label{app:proxy_definitions}

For the proxy comparison in Section~\ref{sec:beyond_importance_similarity}, each fixed sample--layer--source unit has five candidate substitutes. We rank these candidates independently using each proxy and compute its Spearman correlation with the corresponding CFS ranking. Constant ranking rows are omitted, leaving 861 defined ranking units out of 864 valid source units.

For head $i$, the parameter-space vector concatenates its query, key, value, and output-projection parameters,
\begin{equation}
\theta_i
=
\operatorname{vec}
\left(
W_Q^{(i)},W_K^{(i)},W_V^{(i)},W_O^{(i)}
\right),
\end{equation}
excluding biases. Contribution-space comparisons use the flattened post-$W_O$ residual contribution $c_i(x)$. In particular, normalized contribution distance is
\begin{equation}
d_{ij}^{\mathrm{L2}}(x)
=
\frac{
\|c_i(x)-c_j(x)\|_2
}{
\|c_i(x)\|_2+\|c_j(x)\|_2
},
\end{equation}
and contribution cosine distance is
\begin{equation}
d_{ij}^{\cos}(x)
=
1-
\frac{
\langle c_i(x),c_j(x)\rangle
}{
\|c_i(x)\|_2\|c_j(x)\|_2
}.
\end{equation}

The scalar Taylor baseline uses a differentiable head gate,
\begin{equation}
T_i(x)
=
\left|
g_i\frac{\partial\mathcal{L}(x)}{\partial g_i}
\right|_{g_i=1}.
\end{equation}
Taylor therefore measures first-order source importance, whereas CFS measures a directional relation between a source and a candidate substitute.

These controls clarify the interpretation of Table~\ref{tab:cfs_proxy_comparison}: current-space geometric proximity, parameter similarity, and scalar source importance need not identify the alternative path that best reproduces downstream behavior.

\subsection{Additional Input-Conditioning Statistics}
\label{app:input_conditioning}

The input-conditioning analysis uses all 32 frozen inputs and 72 fixed source units from 12 layers and 6 source heads per layer. For each fixed $(\text{layer},\text{source})$ pair, let
\begin{equation}
j_i^*(x)=\arg\max_j S_{i\rightarrow j}(x)
\end{equation}
denote its best substitute.

The median modal-substitute coverage reported in the main text is $31.25\%$. The corresponding median switching rate is therefore $68.75\%$. Across source units, mean modal coverage is $34.07\%$, with first and third quartiles of $28.13\%$ and $37.50\%$. The mean number of distinct best substitutes is $4.97$ out of five candidates, and the median is $5/5$. Thus, the preferred substitute is rarely a fixed identity attached to the source head.

The variation is nevertheless structured. Median best-substitute agreement is $83.3\%$ under mild color perturbations, $50.0\%$ under horizontal flips, and only $16.7\%$ between different images. The full CFS matrices show the same ordering, with median Spearman correlations of $0.955$, $0.870$, and $0.334$, respectively. Functional organization is therefore substantially more stable across transformations of the same input than across unrelated samples, arguing against interpreting input-conditioning as unstructured noise.

\subsection{Directionality}
\label{app:directionality}

For an unordered pair $\{i,j\}$, we quantify directional asymmetry by
\begin{equation}
\Delta_{ij}(x)
=
\left|
S_{i\rightarrow j}(x)
-
S_{j\rightarrow i}(x)
\right|.
\end{equation}
Across 5,760 unordered head pairs, the median absolute directional gap is $0.0785$, and the 75th percentile is $0.1767$. Strongly one-sided relations are less common: $3.94\%$ of pairs have $S>0.10$ in one direction and $S\leq0$ in the reverse, while $1.62\%$ satisfy the stronger condition $S>0.25$ in one direction with a non-positive reverse relation.

These results do not imply that CFS is uniformly highly asymmetric. They instead establish that functional substitutability cannot in general be represented by an undirected similarity relation.

\section{Exact Joint-Subset Oracle: Additional Details}
\label{app:joint_oracle}

Section~\ref{sec:oracle_reduction} evaluates the strongest joint-removal policy available within the measured intervention space. For the final 12-head layer of ViT-B/16, we enumerate
\begin{equation}
\binom{12}{3}=220,
\qquad
\binom{12}{6}=924,
\qquad
\binom{12}{9}=220
\end{equation}
retained subsets for each of 448 frozen test images. Every omitted head is hard-gated simultaneously, and the selected subset is the one with minimum KL divergence from the dense prediction. Across all images and budgets, this produces $611{,}072$ actual joint interventions. Taylor importance is evaluated under exactly the same retained-head budgets and simultaneous-gating protocol.

The KL reductions reported in Table~\ref{tab:joint_oracle} correspond to oracle-minus-Taylor differences of
\begin{equation}
-0.1306,\qquad -0.0384,\qquad -0.0112
\end{equation}
for $K=3,6,9$, respectively. Paired bootstrap resampling over the 448 images gives 95\% confidence intervals
\begin{equation}
[-0.1549,-0.1088],\quad
[-0.0477,-0.0306],\quad
[-0.0154,-0.0079],
\end{equation}
all strictly below zero.

Dense-prediction fidelity also improves at every budget. The corresponding oracle-minus-Taylor fidelity confidence intervals are
\begin{equation}
[0.0089,0.0536],\quad
[0.0223,0.0558],\quad
[0.0000,0.0223].
\end{equation}

The oracle should be interpreted as an upper bound within this intervention space, not as a deployable pruning procedure. Its purpose is to quantify how much collective functional preservation is left unexplained by component-wise importance criteria. Because several removed heads may depend on the same substitute and downstream computation is nonlinear, pairwise CFS does not in general predict the exact quality of a jointly retained subset.

\section{Additional Scaling Protocol and Interpretation}
\label{app:scaling_details}

\subsection{Natural-Family Comparability}
\label{app:natural_scaling_protocol}

The natural scaling experiments are cross-sectional comparisons of pretrained model families rather than single-variable interventions. Their purpose is to test whether a common functional-organization pattern appears across architectures and modalities.

For ViT, the comparison uses Tiny, Small, Base, and Large models on ImageNet-1K. For BERT, the comparison uses pretrained Tiny through Large masked-language models. For Qwen2.5, the main comparison uses the 1.5B, 3B, 7B, and 14B checkpoints. Qwen2.5-0.5B is excluded from the same-granularity scaling sequence because its query heads are 64-dimensional rather than the 128-dimensional query heads used by the larger models.

Raw Oracle and Matched-2 answer complementary questions. Raw Oracle uses every alternative path actually available in a model and therefore reflects both substitute quality and candidate availability. We treat candidate availability as part of the scaled model itself rather than a statistical nuisance: if scaling introduces additional paths that can realize the same downstream role, this increase in functional optionality is part of the model's organization. Matched-2 provides a control in which the candidate count is fixed.

The Matched-2 trajectory is not universally monotonic. ViT follows
\begin{equation}
0.3379,\ 0.2852,\ 0.2889,\ 0.3737,
\end{equation}
whereas same-granularity Qwen2.5 increases from $0.1620$ to $0.2745$ between 1.5B and 14B. Scaling therefore changes not only pairwise substitution strength but also the number of available paths and their global organization. This is why the main claim concerns structural reorganization rather than a universal monotonic trajectory for every individual metric.

\subsection{Controlled Width and Depth Scaling}
\label{app:controlled_scaling_details}

The controlled autoregressive models share the same tokenizer, training corpus, context length, token budget, and optimization protocol. Width scaling fixes depth at 12 layers and head dimension at 64 while using widths $576/768/1152$, corresponding to $9/12/18$ heads. Depth scaling fixes width 768 with 12 heads of dimension 64 and uses $9/12/18$ layers. W768 and D12 are the same model and therefore provide a shared baseline for comparing the two scaling directions.

The shared baseline makes the functional contrast especially clear. Relative to W768/D12, widening to W1152 increases Raw Oracle by $0.0397$, whereas deepening to D18 increases it by only $0.0029$. At the same time, D18 obtains a larger validation-loss reduction per added parameter. Thus, performance improvement does not require proportional growth in available substitutability. The result is consistent with different scaling paths converting added parameters into independent functional structure with different efficiencies.

This comparison should not be read as a general claim that depth is preferable to width. The controlled experiment instead isolates a more limited observation: two capacity increases starting from the same model can achieve different parameter efficiencies while inducing markedly different changes in functional substitutability.

\subsection{Fixed-Capacity Head Decomposition}
\label{app:fixed_capacity_details}

The fixed-capacity experiment holds depth at 12 layers and hidden width at 768 while changing only the decomposition of the attention space into $6\times128$, $12\times64$, or $24\times32$ heads. All three models contain 124.44M parameters and have the same dense attention dimensionality.

Despite matched nominal capacity, the models realize different functional organizations. Validation loss changes from $3.5707$ to $3.5752$ to $3.5780$, while Raw Oracle changes from $0.2012$ to $0.2125$ to $0.2583$ and normalized effective rank from $0.9507$ to $0.9335$ to $0.8725$. The model with lower substitutability and greater normalized independent structure achieves the lower loss.

Because changing head granularity also changes architectural inductive bias, this experiment establishes association rather than a causal effect of directly optimizing CFS. Its role is to show that parameter count alone does not uniquely determine functional organization: architectures with identical nominal capacity can realize different amounts and arrangements of independent functional structure.

\section{Additional Routing Details and Ablations}
\label{app:routing_details}

\subsection{R1 and R2 Supervision}
\label{app:routing_supervision}

The routing experiments use a frozen ViT-B AugReg backbone. The router is trained on 4,096 images, tuned on 64 disjoint images, and evaluated on 448 held-out images using seeds 17, 23, and 41. The dense backbone achieves $81.92\%$ accuracy.

R1 combines pairwise CFS ranking and regression, source weighting, pairwise compensation prediction, and structured subset supervision. R2 retains these objectives and adds direct joint-subset supervision to address composition errors that cannot be inferred from pairwise CFS alone.

For each training image and retained-head budget, the joint teacher evaluates 48 candidate masks by simultaneous hard gating and records their KL divergence from the dense model. Candidate masks include CFS- and router-based anchors, Taylor and magnitude baselines, similarity-based subsets, hard negatives, and random subsets. R2 is trained to align predicted subset utilities with these joint intervention outcomes.

At inference time, neither pairwise interventions nor teacher masks are available. The router receives only the pre-attention residual summary defined in Section~\ref{sec:cfs_routing}, and the backbone remains frozen. For 12 heads, subset selection is exact rather than greedy, enumerating $220/924/220$ subsets at $K=3/6/9$.

\subsection{Routing Confidence Intervals}
\label{app:routing_ci}

The paired 95\% confidence intervals for CFS-minus-Taylor KL at $K=3,6,9$ are
\begin{equation}
[-0.1137,-0.0717],\quad
[-0.0325,-0.0157],\quad
[-0.0110,-0.0035],
\end{equation}
respectively. All are strictly below zero. Accuracy is directionally higher under CFS at all three budgets, but its confidence intervals overlap zero; we therefore use dense-output fidelity as the primary routing evidence.

To separate functional subset selection from compensation quality, we also compare CFS- and Taylor-selected subsets under shared calibration. With the same learned pairwise calibration, the CFS KL advantages are
\begin{equation}
0.0696,\quad 0.0178,\quad 0.0068
\end{equation}
for $K=3,6,9$. When both methods instead receive oracle pairwise compensation, the corresponding advantages remain
\begin{equation}
0.0604,\quad 0.0154,\quad 0.0049.
\end{equation}
All six paired confidence intervals exclude zero. The routing improvement therefore cannot be explained by compensation alone; the functional relations lead to better subset selection.

\subsection{Joint-Subset Supervision Ablation}
\label{app:routing_r2}

Relative to R1, R2 changes KL from
\begin{equation}
0.1217/0.0315/0.00874
\end{equation}
to
\begin{equation}
0.1190/0.0281/0.00699
\end{equation}
at $K=3/6/9$. The improvement is statistically resolved at $K=6$ and $K=9$, while the $K=3$ gain is borderline. This supports the motivation for joint supervision: pairwise CFS captures useful relational structure, but simultaneous removal introduces higher-order interactions that pairwise measurements alone do not fully specify.

\subsection{Online Execution Check and Systems Boundary}
\label{app:routing_execution}

When the predicted masks are integrated into the forward pass, online and frozen offline masks agree on $447/448$ test images, or $99.78\%$, at every budget. This verifies that the reported routing policy is reproduced by the online implementation.

The current implementation should nevertheless be interpreted as logical component allocation rather than realized wall-clock acceleration. The backend still executes fused dense QKV and output-projection kernels, so masking heads does not directly translate into proportional runtime savings. Real systems gains would require an execution backend capable of exploiting input-dependent structured sparsity.

\end{document}